\documentclass[letterpaper, 10 pt, conference]{IEEE_styles/ieeeconf}  

\IEEEoverridecommandlockouts 
\usepackage{0a_preamble}
\usepackage{todonotes}

\begin{document}

\setlength{\abovedisplayskip}{3pt}
\setlength{\belowdisplayskip}{3pt}


\newcommand\ProjectName{RadCloud}

\newcommand\radar{radar}
\newcommand\lidar{lidar}

\newcommand\iwr{TI-IWR1443}
\newcommand\dca{TI-DCA1000}

\newcommand{\sect}[1]{Sec.~{#1}}
\newcommand{\fig}[1]{Fig.~{#1}}
\newcommand{\eq}[1]{eq.~{#1}}
\newcommand{\tbl}[1]{Table~{#1}}

\newcommand{\dBsm}[1]{{#1}\thinspace{dBsm}}
\newcommand{\dBsmSq}[1]{{#1}\thinspace{$\textrm{dBsm}^2$}}
\newcommand{\dB}[1]{{#1}\thinspace{dB}}
\newcommand{\dBm}[1]{{#1}\thinspace{dBm}}
\newcommand{\nsec}[1]{{#1}\thinspace{ns}}
\newcommand{\usec}[1]{{#1}\thinspace{$\upmu$s}}
\newcommand{\msec}[1]{{#1}\thinspace{ms}}
\newcommand{\seconds}[1]{{#1}\thinspace{s}}
\newcommand{\GHz}[1]{{#1}\thinspace{GHz}}
\newcommand{\MHz}[1]{{#1}\thinspace{MHz}}
\newcommand{\kHz}[1]{{#1}\thinspace{kHz}}
\newcommand{\Hz}[1]{{#1}\thinspace{Hz}}
\newcommand{\MSps}[1]{{#1}\thinspace{MSa/s}}
\newcommand{\MBps}[1]{{#1}\thinspace{MBps}}
\newcommand{\Mbps}[1]{{#1}\thinspace{Mbps}}
\newcommand{\Gbps}[1]{{#1}\thinspace{Gbps}}

\newcommand{\MHzPerus}[1]{{#1}\thinspace{MHz/$\upmu$s}}
\newcommand{\m}[1]{{#1}\thinspace{m}}
\newcommand{\cm}[1]{{#1}\thinspace{cm}}
\newcommand{\mm}[1]{{#1}\thinspace{mm}}
\newcommand{\mPers}[1]{{#1}\thinspace{m/s}}

\newcommand{\degrees}[1]{${#1}^{\circ}$}

\newcommand{\grams}[1]{{#1}\thinspace{g}}

\newcommand{\Watts}[1]{{#1}\thinspace{W}}

\newcommand{\LightSpeed}{\textrm{c}}


\newcommand{\NumChirpsPerFrame}{N_{\textrm{chirps}}}
\newcommand{\FrameDuration}{T_{\textrm{frame}}}
\newcommand{\ChirpDuration}{T_{\textrm{chirp}}}
\newcommand{\ChirpFreqStart}{f_{c}}
\newcommand{\ChirpSlope}{S}
\newcommand{\ChirpSlopeVic}{S_{\textrm{vic}}}
\newcommand{\ChirpSlopeAtt}{S_{\textrm{atk}}}
\newcommand{\ChirpBW}{B}
\newcommand{\FreqIF}{f_{\textrm{IF}}}
\newcommand{\FreqSamp}{f_{\textrm{samp}}}
\newcommand{\NSamps}{N_{\textrm{Samp}}}
\newcommand{\NChirps}{N_{\textrm{Chirps}}}
\newcommand{\NRx}{N_{\textrm{Rx}}}

\newcommand{\RangeRes}{d_{\textrm{res}}}
\newcommand{\RangeMax}{d_{\textrm{max}}}
\newcommand{\RangeMin}{d_{\textrm{min}}}
\newcommand{\VelocityRes}{v_{\textrm{res}}}
\newcommand{\VelocityMax}{v_{\textrm{max}}}
\newcommand{\AngleRes}{\theta_{\textrm{res}}}
\newcommand{\DopplerShift}{\phi_{\textrm{doppler}}}
\newcommand{\tDelay}{t_{\textrm{d}}}
\newif\ifShowComments
\newif\ifShowToDos

\ShowCommentstrue
\ShowToDostrue

\newcommand\tingjun[1]{%
    \ifShowComments
        \textcolor{red}{[TC: #1]} 
    \fi%
}

\newcommand\david[1]{
    \ifShowComments\textcolor{blue}{[DH: #1]} \fi%
}

\newcommand\newText[1]{
    \textcolor{blue}{#1}
}

\newcommand\shaocheng[1]{
    \ifShowComments\textcolor{brown}{[SL: #1]} \fi%
}

\newcommand\xiao[1]{
    \ifShowToDos\textcolor{green}{[Xiao: #1]}\fi%
}

\title{\LARGE \bf 
\ProjectName: 
Real-Time High-Resolution Point Cloud Generation \\ Using Low-Cost Radars for Aerial and Ground Vehicles
}

\newif\ifAnonymize

\Anonymizefalse

\ifAnonymize

\else
    \author{David Hunt, Shaocheng Luo, Amir Khazraei, Xiao Zhang, Spencer Hallyburton, Tingjun Chen, Miroslav Pajic %
    \thanks{© 2024 IEEE.  Personal use of this material is permitted.  Permission from IEEE must be obtained for all other uses, in any current or future media, including reprinting/republishing this material for advertising or promotional purposes, creating new collective works, for resale or redistribution to servers or lists, or reuse of any copyrighted component of this work in other works.}
    \thanks{*This work is sponsored in part by the ONR N00014-23-1-2206  and AFOSR FA9550-19-1-0169 awards, NSF  CNS-1652544, CNS-2211944, and AST-2232458 grants, and the National AI Institute for Edge Computing Leveraging Next Generation Wireless Networks, Grant CNS-2112562.}
    \thanks{The authors are with the Department of Electrical and Computer
    Engineering, Duke University, Durham, NC 27708 USA (e-mail:
    \{david.hunt, shaocheng.luo, xiao.zhang, amir.khazraei, spencer.hallyburton, tingjun.chen, miroslav.pajic\}@duke.edu).}
    }
\fi

\maketitle

\thispagestyle{empty}
\pagestyle{empty}

\begin{abstract}
In this work, we present {\ProjectName}, a novel \emph{real-time} framework for 
directly obtaining
higher-resolution {\lidar}-like 2D point clouds from 
low-resolution {\radar} frames 
on
resource-constrained platforms commonly used in unmanned aerial and ground vehicles (UAVs and UGVs, respectively); such point clouds can then be used for accurate environmental mapping, navigating unknown environments, and other 
robotics tasks. 
While 
high-resolution sensing using \radar\ data has been previously reported, 
existing methods cannot be used on most UAVs, which have limited computational power and energy;  
thus, existing demonstrations focus on offline {\radar} processing.
{\ProjectName} overcomes these challenges by using a {\radar} configuration with 1/4th of the range resolution and employing a deep learning model with $2.25\times$ fewer parameters. 
Additionally, {\ProjectName} utilizes a novel chirp-based approach 
that makes 
obtained point clouds resilient to rapid movements (e.g., aggressive turns or spins) that commonly occur during UAV flights. 
In real-world experiments, we demonstrate~
the accuracy and applicability of {\ProjectName} on commercially available UAVs and UGVs, with off-the-shelf radar platforms on-board.
\end{abstract}

\section{Introduction}
\label{sec:introduction}

Light detection and ranging (\lidar) sensors are often referred to as the golden standard for applications requiring highly accurate and dense 3D point clouds~\cite{dong_lidar_2017}. For example, the common Velodyne VLP-16 Puck has a horizontal angular resolution of \degrees{0.1} and a range resolution of \mm{2}~\cite{velodyne_lidar_vlp-16_2018}. Such high ranging and angular resolutions make {\lidar} sensors particularly well suited for various applications including mapping, navigation, surveying, and advanced driver assistance systems~\cite{royo_overview_2019,payne_autonomous_nodate}. However, these sensors also have poor performance in low-visibility environments like fog and smoke. Additional drawbacks include high cost, large form factors (i.e., size), and higher power consumption compared to other ranging sensors on the market. For example, the VLP-16 {\lidar} requires a separate interface box, consumes \Watts{8} of power during nominal operation, has a mass of \grams{830}, and costs \$4,600~\cite{velodyne_lidar_vlp-16_2018,noauthor_puck_nodate}, with drone-mounted sensors costing even more~\cite{noauthor_zenmuse_nodate}. Thus, most {\lidar}s are
ill-suited for resource-constrained vehicles, such as small to midsize~UAVs. 

On the other hand, millimeter-wave (mmWave) radio~detection and ranging (\radar) sensors are cheaper, smaller, lighter, and consume far less power while also providing accurate ranging information even in adverse weather and lighting conditions~\cite{keysight_how_2020,benjamin_imaging_2019,gardill_automotive_2019}. For example, the commercially available {\iwr} mmWave {\radar} sensor has a typical power consumption of \Watts{2}, weighs \grams{245}, and the full evaluation kit costs only \$398~\cite{mouser_iwr1443boost_nodate, texas_instruments_iwr1443_2018}. 
Due to the large signal bandwidth of up to \GHz{4}, mmWave {\radar}s 
can achieve cm-level range resolutions (e.g., \cm{4} for {\iwr}~\cite{keysight_how_2020,ramasubramanian_moving_2017}). 

However, {\radar}s suffer from poor angular resolution. For example, the \iwr\ has a maximum azimuth~resolution of \degrees{30} \cite{rao_introduction_nodate}. Thus, our goal is to enable real-time, affordable, and high-resolution sensing on resource constrained vehicles (e.g., UAVs) by using 
deep learning 
to overcome the traditional resolution limits of \radar\ sensors. 

In particular, this work introduces {\ProjectName}, a novel \emph{real-time} framework for efficient generation of high-resolution {\lidar}-like 2D point clouds using low-resolution {\radar} data for resource-constrained unmanned vehicles (e.g., UAVs and UGVs). While several recent works have explored high-resolution sensing and mapping applications using mmWave \radar\ sensors~\cite{avidan_radatron_2022-1, meyer_graph_2021, qian_3d_2020, yanik_development_2020, gao_mimo-sar_2021, sengupta_review_2020, schreiber_advanced_2019, qian_3d_2020, lu_see_2020, cai_millipcd_2022},
they all have limitations such as only working on specific applications, relying on highly accurate position information (e.g.,~\cite{qian_3d_2020,yanik_development_2020, gao_mimo-sar_2021,sengupta_review_2020, schreiber_advanced_2019}) or not working in real-time (i.e., using offline processing). {To start, \cite{Cheng_2022_Novel} utilized a deep learning model to better detect real objects and filter out false {\radar} defections to achieve higher quality point clouds. Also, \cite{avidan_radatron_2022-1,meyer_graph_2021} focus on generating accurate 3D bounding boxes from \radar\ data, but can only identify specific objects (e.g., vehicles) in the environment. By contrast, \ProjectName\ generates a \lidar-like point cloud from raw {\radar} data for the \emph{entire} environment, including stationary objects like walls.} Similarly, \ProjectName\ does not require position information and allows the sensing platform (i.e., the vehicle) to follow any trajectory. 

Outside of traditional radar processing methods, \cite{qian_3d_2020,lu_see_2020,cai_millipcd_2022} present methods of converting \radar\ data into \lidar-like point clouds for the purposes of indoor mapping. While \cite{qian_3d_2020,lu_see_2020,cai_millipcd_2022} post-process \radar\ scans of a particular scene or indoor environment to create final \lidar-like point cloud mapping, \ProjectName\ directly converts \radar\ data frames into 2D \lidar-like point clouds in \emph{real-time}. Thus, enabling the use of the generated point clouds for other purposes like real-time navigation (in addition to mapping) and SLAM.

To the best of our knowledge, only \cite{prabhakara_high_2023} recently presented a method of directly converting raw \radar\ data into \lidar-like point clouds using a modified U-Net architecture~\cite{ronneberger_u-net_2015}. However, 
the previously recorded radar data, sampled at the highest resolution for the \radar\ sensor, were processed offline. On the other hand, we experimentally discovered~that common UAV compute platforms (e.g., Intel NUC) cannot process the raw data from the \radar\ at sufficient rates to support the highest resolution \radar\ configurations from~\cite{prabhakara_high_2023}. Hence, {\ProjectName} utilizes radar data with 1/4th the maximum possible range resolution. Moreover, the use of lower-resolution radar data allows for the use of a model with less depth and 2.25× fewer parameters than the model from~\cite{prabhakara_high_2023}, reducing computational overhead. Despite these constraints due to the real-time radar processing, 
we show that {\ProjectName}  model generates sufficiently accurate (non-inferior to~\cite{prabhakara_high_2023}) 2D \lidar-like point clouds with 90\% of predictions having errors $<$\cm{40} compared to the ground truth \lidar ~data.

Additionally, \cite{prabhakara_high_2023} utilized the 40 most recent single-chirp frames to improve the accuracy of the generated point~cloud. Yet, even with high {\radar} sampling rates and offline processing, we find that this approach becomes significantly less accurate during rapid changes in orientation or direction (e.g., rapid vehicle turning or spinning). Thus, in contrast to such frame-based approach using the previous \seconds{2} of sensing data, we utilize 40 radar chirps collected over a period of \msec{8}. In real-world experiments, deploying {\ProjectName} on a UAV and a UGV, we demonstrate that this chirp-based~approach is much more resilient to aggressive maneuvers. To the best of our knowledge, this is the first work to implement a completely real-time framework for directly converting \radar\ data into 2D \lidar-like point clouds on resource-constrained vehicles.

This paper is organized as follows. \sect{\ref{sec:background}} overviews the employed radar signal processing pipeline. \sect{\ref{sec:system_design}} describes the system model, starting with the \radar\ setup, before introducing the deep learning model used to generate the high-resolution point clouds. \sect{\ref{sec:experiments}} presents the experimental setup for {\ProjectName} ~evaluation, followed by evaluation results (\sect{\ref{sec:results}}), and concluding remarks (\sect{\ref{sec:conclusion}}). {Additional resources, including code and datasets are available at \cite{RadCloud_Website}.}

\section{Background: Radar Signal Processing}
\label{sec:background}

\begin{figure}[!t]
\centering
\includegraphics[width=0.936\columnwidth]{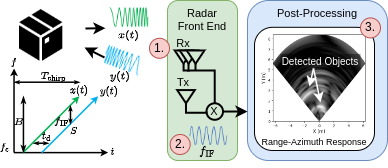}
\vspace{-6pt}
\caption{Radar Signal Processing Pipeline.}
\vspace{-6pt}
\label{fig:fmcw_overview}
\end{figure}

\begin{figure*}[!t]
\centering
\includegraphics[width=1.94\columnwidth]{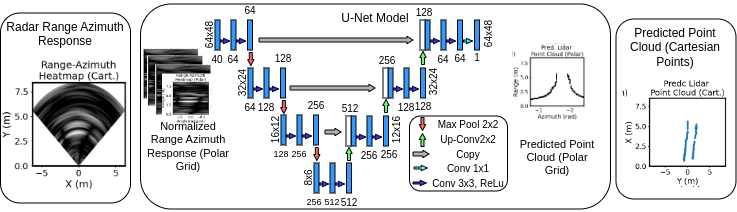} 
\vspace{-6pt}
\caption{\ProjectName\ framework overview.}
\vspace{-6pt}
\label{fig:framework_overview}
\end{figure*}
We consider a frequency-modulated continuous wave (FMCW) \radar\ sensor. Here, the \radar's transmitter (Tx) transmits a specifically constructed signal into the environment. The signal reflects off objects which are then received by the \radar's receiver (Rx). {While {\radar} sensors can detect an object's range, velocity, and angle, in this initial work we only focus on the range and angle information.}  
%
The pipeline we employ is composed of three steps~(\fig{\ref{fig:fmcw_overview}}).

\vspace{4pt}
\noindent\textbf{Step \circled{1}: Tx and Rx chirps}. For each frame, the \radar\ transmits a series of ``\emph{chirps}'' whose frequency increases linearly over time. In our framework, we transmit up to 40 chirps per frame (see \sect{\ref{sec:system_design}}). Here, we use the following notation: $\ChirpSlope$ denotes the \emph{chirp slope}, $\ChirpFreqStart$ denotes the \emph{chirp start frequency}, and $\FreqIF$ denotes the \emph{intermediate frequency} (IF) from mixing the Tx and Rx signals. Thus, the Tx signal for a single chirp in the \radar\ frame is given by~\cite{alland_interference_2019,jiang_mmvib_2020, wang_remote_2020}
\begin{equation}
\label{eq:FMCW_tx_chirp}
x(t) = e^{ j \left(2 \pi \ChirpFreqStart \cdot t + \pi \ChirpSlope \cdot t^2 \right) }.
\end{equation}
If we assume that the environment is stationary, the time that it takes for the transmitted signal to propagate to a target at range of $d$ and back at the speed of light ($\LightSpeed$)  is given by $\tDelay = 2 d/\LightSpeed$. Thus, the received signal can be captured as
\begin{equation}
\label{eq:FMCW_rx_chirp}
    y(t) = A_{\textrm{Rx}} \cdot e^{ j \left[2 \pi \ChirpFreqStart (t-\tDelay) + \pi \ChirpSlope (t - \tDelay)^2 \right] },
\end{equation}
where $A_{\textrm{Rx}}$ denotes the received signal amplitude.

\vspace{4pt}
\noindent\textbf{Step \circled{2}: Dechirping and IF signal generation}.
Next, the IF signal is obtained by mixing the Tx and Rx signals
\begin{equation}
\label{eq:FMCW_IF_signal_simplified}
    s_{\textrm{IF}}(t) = x(t) \cdot y^{*}(t)
    = A_{\textrm{IF}} \cdot e^{j 2 \pi \FreqIF \cdot t} \c,
\end{equation}
where $\FreqIF := \frac{2 \ChirpSlope \cdot d}{c}$, $\lambda = \LightSpeed/\ChirpFreqStart$ is the signal wavelength, and $A_{IF}$ is the amplitude of the IF signal~\cite{wang_remote_2020}. Then, the IF signal is put through a low-pass filter (typically removing all IF frequencies above \MHz{10--20}) and sampled by an analog-to-digital converter (ADC) at a rate $\FreqSamp$. For each chirp, a total of $\NSamps$ I-Q samples are recorded.

\vspace{4pt}
\noindent\textbf{Step \circled{3}: Range-Azimuth response}.
Using a fast Fourier transform (FFT), commonly called the ``RangeFFT'', the IF frequency corresponding to a target is estimated; the target's range is determined using $d_{object} = \frac{\FreqIF}{2\ChirpSlope} \cdot c.$ The range resolution ($\RangeRes$ -- the minimum distance between two objects detectable by a {\radar}) and maximum range ($\RangeMax$) are \cite{rao_introduction_nodate}
\begin{equation} \label{eq:range_performance}
    \RangeRes = \frac{\LightSpeed}{2 \ChirpBW},\hspace{10pt}
    \RangeMax = \frac{\FreqSamp \cdot \LightSpeed}{\ChirpSlope}.
\end{equation}

{mmWave {\radar}s use multiple receive elements to determine the angle of an object in the environment. 
The angular resolution for objects at boresight is defined as $\AngleRes = 2/\NRx$ (radians), where $\NRx$ is the number of Rx antennas~\cite{rao_introduction_nodate}. By sampling the IF signal, a 2D FFT 
can be used to compute a Range-azimuth response (\fig{\ref{fig:fmcw_overview}}); e.g., the {\iwr} {\radar} with 4 Rx elements achieves a maximum $\AngleRes$ of \degrees{28.6}.}

\section{\ProjectName\  Design}
\label{sec:system_design}

We designed \ProjectName\ to operate completely in real-time while also being robust to rapid movements commonly experienced by UAVs and UGVs. These design goals impacted our system design in several ways.
\subsection{Radar Setup}
Unlike all previous work, which employs offline processing of highest resolution radar\ data,
our system processes the raw sensor data in real-time. This is an important distinction as most UAV platforms do not have a high enough computational bandwidth to support the instantaneous data rates required to process {\radar} sensor data at the highest~resolution.

We use the {\iwr} {\radar} sensor to perform sensing, and the \dca\ data capture card to send the raw data to the host~\cite{texas_instruments_iwr1443_2018,texas_instruments_iwr1443_2020,texas_instruments_dca1000evm_2019}. This platform has been used in most previous works involving mmWave {\radar} (e.g.,~\cite{qian_3d_2020,lu_see_2020,cai_millipcd_2022,prabhakara_high_2023,hunt_ndss24}). Here, the {\radar} sends complex-valued samples of the IF signal, $s_{\textrm{IF}}(t)$, captured at rate $\FreqSamp$,~where each sample is 4-Byte (16-bit integer for real and 16-bit integer for complex). Thus, operating the {\radar} at the maxi- mum $\FreqSamp$ of \MSps{18.75} requires the platforms to support an instantaneous data rate of \Gbps{2.4}~\cite{texas_instruments_iwr1443_2018, texas_instruments_iwr1443_2020, texas_instruments_dca1000evm_2019}.

Unfortunately, many UAV platforms do not have sufficiently high computation bandwidth to support data sent 
at such high instantaneous rates due to limited computational resources. Also, \radar\ configurations with $\NSamps \geq 90$ require multiple packets per chirp (due to the maximum Ethernet packet size of 1462 B~\cite{texas_instruments_dca1000evm_2019}). For our platform, we empirically decided to operate the \radar\ with $\FreqSamp=$ \MSps{2} (instantaneous data rate of \Mbps{256}) and $\NSamps=64$\footnote{{Empirical observations showed a significant number of dropped Ethernet packets for configurations requiring multiple Ethernet packets per chirp.}} to ensure real-time data processing and avoid~packet~losses.

\vspace{4pt}
\noindent\textbf{Radar Configuration.} Given the constraints imposed by the considered UAV platforms, we selected a \radar\ configuration that maximized the range resolution ($\RangeRes$) while achieving a maximum range ($\RangeMax$) of roughly \m{10}. Thus, our final configuration utilized chirps with $\ChirpSlope =$\MHzPerus{35}, $\FreqSamp =$\MSps{2}, and $\NSamps=64$, achieving a chirp bandwidth ($\ChirpBW$) of \GHz{1.12}. From~\eqref{eq:range_performance}, we achieve a $\RangeRes=$ \cm{13.3} and $\RangeMax=$\m{8.56}. Compared with previous works (e.g.,~\cite{prabhakara_high_2023} that used the maximum $\ChirpBW=$ \GHz{4} with $\NSamps=256$ to achieve $\RangeRes=$ \cm{3.7} and $\RangeMax=$ \m{9.59}), the real-time data streaming constraints restrict our system to utilizing \radar\ data with roughly 1/4th the range resolution. {Finally, we note the  {\radar} configuration only performs 2D sensing in light of the sensor's {6}\thinspace{dB} elevation beamwidth of $\pm$\degrees{20}.}

\subsection{Deep Learning (DL) Model}
We develop a DL model based on the U-Net architecture~\cite{ronneberger_u-net_2015}, which takes in a normalized Range-Azimuth response from the {\radar} (in polar coordinates) and outputs a quantized grid representing the {\lidar}-like point cloud (in polar coordinates). The output point cloud is then converted into the 2D cartesian coordinates that can be used for navigation, mapping, or other point cloud detection algorithms.
\fig{\ref{fig:framework_overview}} presents an overview of the {\ProjectName} framework.

\vspace{4pt}
\noindent\textbf{Model Input.} 
For each {\radar} chirp, we compute a complex-valued range-azimuth response (Step 3 from \sect{\ref{sec:background}}) with 64 range bins and 64 azimuth bins\footnote{{The 64 azimuth bins are achieved by zero-padding the azimuth FFT's input to 64 bins to increase the smoothness of the generated response. The resulting azimuth FFT has an FFT bin resolution (different from $\AngleRes$) of \degrees{1.8} at boresight and \degrees{15} at \degrees{90} off of boresight~\cite{rao_introduction_nodate,texas_instruments_iwr1443_2020}.}}. To convert the response into a format that can be used by a DL model, we start by taking the magnitude of the complex data, applying a threshold to filter out very weak reflections\footnote{We filter out all reflections that are \dB{45} less than the maximum reflected signal power as such reflections are often just noise.}, and normalize the response to between 0 and 1. While the field of view of the radar is theoretically $\pm$\degrees{90} due to the Rx element spacing, we only use parts of the Range-Azimuth response that are within $\pm$\degrees{50} as this is the horizontal \dB{6} beamwidth of the {\radar}, and the radar's angular resolution significantly decreases at angles outside of this field-of-view~\cite{rao_introduction_nodate,texas_instruments_iwr1443_2020}.

Compared to \cite{prabhakara_high_2023}, which utilized \seconds{2} worth of previous frames to improve model accuracy, we empirically decided to employ 40 chirps from a single radar frame, requiring only \msec{8} of total {\radar} sensing time, i.e., a  250$\times$ reduction. Thus, the final input to the model is a 40$\times$64$\times$48\footnote{The reduction of the azimuth dimmension from 64 to 48 occurs due to the narrowing the \radar's FOV from $\pm$ \degrees{90} to $\pm$\degrees{50}.} tensor corresponding to the 40 normalized range azimuth responses. 

Our chirp-based approach is particularly important because a \radar's view of the environment changes rapidly when the platform experiences rapid movement (e.g., a vehicle spinning/turning quickly or moving at high speeds). For the previous frame-based approach~\cite{prabhakara_high_2023}, this causes the scene captured in the first few frames to vary drastically compared to the scene captured in the last few frames. As we show in \sect{\ref{sec:results}}, the performance of these frame-based models noticeably degrades when moving rapidly because the sensed environment can change significantly over several frames in such cases. Thus, we show that our chirp-based approach \emph{significantly improves robustness to aggressive maneuvers}.

\vspace{4pt}
\noindent\textbf{Model Output.} 
We take three steps when pre-processing the ground-truth {\lidar} data used to train and evaluate our model's performance. To start, we filter the {\lidar} point cloud so that it has the same azimuth field of view as the input radar data. Next, we obtain a 2D {\lidar} point cloud by only keeping points with  \cm{$-$20} $\leq$ z $\leq$ \cm{10}  to filter out undesired ground detections and in light of the {\radar} sensor's 2D configuration.
Here, we match the input and output dimensions by converting the Cartesian point cloud to polar coordinates and then quantizing the points into a 64$\times$48 polar grid with a horizontal resolution of \degrees{2.08} and a range resolution of \cm{13.3}. Thus, to obtain a {\lidar}-like point cloud from the model's prediction, we convert the prediction grid into a set of Cartesian points.

\vspace{4pt}
\noindent\textbf{Model Architecture.} 
We use a simplified U-net architecture~\cite{ronneberger_u-net_2015} to generate the higher resolution point clouds. The primary benefit of this architecture is its encoder-decoder structure. In our case, the encoder progressively downsamples the input data while capturing key context and feature information from the input radar data. Then, the decoder outputs the higher resolution point cloud by generating features from the encoded data while also preserving spatial information through the use of skip connections ~\cite{ronneberger_u-net_2015}. 

The final model architecture is shown in \fig{\ref{fig:framework_overview}}. To ensure our model's real-time performance on resource constrained platforms, we implemented a simplified model architecture with fewer layers and less depth compared to the model from~\cite{prabhakara_high_2023};
our model utilizes $\sim${7.7}\thinspace{M} parameters compared to the  $\sim${17.5}\thinspace{M} parameters used by~\cite{prabhakara_high_2023}, demonstrating a 2.25$\times$ reduction in the number of parameters. As we show in the following sections, this simpler model allows us to achieve real-time frame rates, even on CPU-only machines. 

\vspace{4pt}
\noindent\textbf{Loss Function.}
As shown in~\cite{prabhakara_high_2023}, utilizing a combination of Binary Cross Entropy (BCE) loss and Dice loss 
is particularly effective when converting {\radar} data to {\lidar}-like point clouds. Here, the BCE loss seeks to force each predicted pixel to be as close to the actual value as possible while the added Dice loss helps to make the predicted features sharper. Similar to~\cite{prabhakara_high_2023}, we weighted the BCE loss to be 0.9 while weighting the Dice loss by a factor of 0.1.

\section{Experiments}
\label{sec:experiments}

\subsection{Experimental Platform}
\begin{figure}[!t]
\centering
\includegraphics[width=0.84\columnwidth]{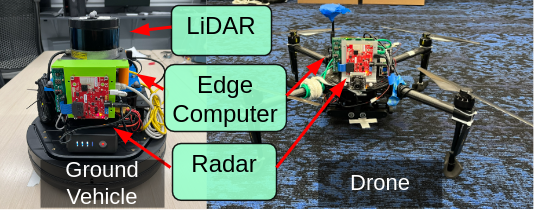}
\vspace{-6pt}
\caption{UGV and UAV experimental platforms.}
\vspace{-6pt}
\label{fig:experimental_implementation}
\end{figure}
We demonstrate the real-world capability of \ProjectName\ by implementing 
it on a common 
vehicle platform. 

\vspace{4pt}
\noindent\textbf{Radar.} 
We utilize the commercially available \iwr\ \radar\ to sense the environment and the \dca\ data capture card to stream the raw \radar\ data to the edge compute platform in real-time ~\cite{texas_instruments_iwr1443_2018,texas_instruments_dca1000evm_2019,texas_instruments_iwr1443_2020}. Here, we implemented a ROS-compatible real-time streaming interface in Python to obtain data from the \iwr\ via the \dca. 

\vspace{4pt}
\noindent\textbf{Lidar.} 
To obtain ground truth information, we use the popular VLP-16 Puck \lidar\ sensor ~\cite{velodyne_lidar_vlp-16_2018,noauthor_puck_nodate}. {Here, we ensure a consistent extrinsic calibration between the {\radar} and {\lidar} sensors by mounting the {\radar} sensor \cm{7} below and \cm{10} in front of the {\lidar} sensor, enabling the model to `learn' the relative position of the {\radar} with respect to the {\lidar}.}

\vspace{4pt}
\noindent\textbf{Vehicle Computer}. We implement our entire real-time framework on the NUC7i5BNH  as our vehicle computing platform~\cite{intel_intel_2023}. Unlike previous works (e.g.,~\cite{prabhakara_high_2023}), which~use  powerful Jetson platforms with an integrated GPU, our~platform only has a dual-core \GHz{3.4} Intel i5 CPU and no GPU. \emph{We highlight that we are the first to implement a completely real-time system, including streaming data from \radar\ to the NUC7i5BNH, range-azimuth response computation, and generating high-resolution 2D \lidar-like point clouds.} While we operate our system at 10 frames per second, we achieved average frame rates above 15 frames per second when testing our full pipeline on the NUC7i5BNH.
Overall, the full ROS-compatible framework is implemented in over 7,000 lines of code,  which is responsible for the real-time capturing of {\radar} data and its conversion into {\lidar}-like 2D point clouds.

\vspace{4pt}
\noindent\textbf{Robotic Platforms.}
We mount the
\ProjectName~platform on a Kobuki ground vehicle and a DJI Matrice 100 drone~\cite{kobuki_kobuki-userguidepdf_2016, dji_matrice_2023} (\fig{\ref{fig:experimental_implementation}}). While we were able to mount the VLP-16 {\lidar} onto the ground vehicle, we were unable to mount it on the drone due to the size and weight limitations on the drone. Thus, we utilize the ground vehicle platform to assess model performance and the drone-based platform to demonstrate the feasibility of our framework in airborne environments.

\subsection{Experimental Setup}

\begin{figure}[!t]
\centering
\includegraphics[width=0.646\columnwidth]{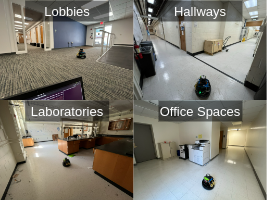} 
\vspace{-6pt}
\caption{Training and Testing Environments.}
\label{fig:training_environments}
\vspace{-6pt}
\end{figure}

\noindent\textbf{Training, Validation, and Test Datasets.}
For training, validation, and initial testing of the {\ProjectName} model, we capture time-synchronized \radar~and \lidar~frames, sampled at a frame rate of \Hz{10}, across 7 different environments including laboratories, corridors, and building lobbies (\fig{\ref{fig:training_environments}}). To improve model robustness, we also drive the ground~vehicle along various trajectories including turns, spinning, and straight-line movement at a range of angular and linear velocities. We recorded a total of 87,476 samples for training, validation, and testing; we used 66,609 samples for training, 11,755 samples for validation and parameter tuning, and 9,112 samples for testing. The test dataset was recorded independently from the training and validation datasets, and it included unique trajectories. Thus, we were able to assess our model's performance when operating in the ``same'' environment.

\vspace{4pt}
\noindent\textbf{Unseen Environment Test Dataset.}
In addition to the 7 environments used for training, validation, and initial testing, we also recorded an additional 4,767 samples in 3 unseen environments that the model had not previously been trained on. We used the results of this ``unseen environment'' dataset to assess the model's performance when operating in unfamiliar environments, similar to how a UAV or a UGV may be used to map or navigate in unknown environments.

\vspace{4pt}
\noindent\textbf{Rapid Movement Test Dataset.} Finally, we record a third dataset specifically to evaluate our model's resiliency to rapid movements (e.g., spinning, fast turns, and high speed movements) commonly encountered by UAVs and UAGs. While both the training and test set do include some aggressive movements, the majority of \radar\ and \lidar\ frames were recorded at relatively slower speeds. Thus, we recorded an additional 784 samples while driving the UGV along different trajectories at maximum speed and rotational velocity, to enable evaluating the model's performance in such situations.

\subsection{Evaluation Metrics}
We utilize the commonly used Chamfer and Modified Hausdorff metrics to evaluate the accuracy of {\ProjectName} model's predicted point cloud compared to the ground truth point cloud obtained from the \lidar~\cite{bell_chamfer_2023,watkins_chamfer_2023,dubuisson_modified_1994}. Here, we define the Chamfer distance (CD) as
\begin{equation}
\label{eq:chamfer_distance}
    \begin{split}
       \textrm{CD}(S_1,S_2) = &
            \frac{1}{2|S_1|}\underset{{x\in S_1}}{\Sigma} \underset{y \in S_2}{\textrm{min}}d(x,y)
             + \frac{1}{2|S_2|}\underset{{y\in S_2}}{\Sigma} \underset{x \in S_1}{\textrm{min}}d(x,y),
    \end{split}        
\end{equation}
and Modified Hausdorff distance (MHD) as
\begin{equation*}
\label{eq:mod_haus_distance}
        \textrm{MHD}(S_1,S_2)  =
             \textrm{max} \left\{ \underset{{x\in S_1}}{\textrm{med}} \underset{y \in S_2}{\textrm{min}}d(x,y),  
            \underset{{y\in S_2}}{\textrm{med}} \underset{x \in S_1}{\textrm{min}}d(x,y)\right\},
\end{equation*}
where $d(x,y)$ denotes the Euclidean distance i.e., $||x -   y||_2^2$.

\subsection{Comparison to Baseline}
We were unable to compare our model's performance with the model from~\cite{prabhakara_high_2023} as the input and output data dimensions for {\ProjectName} model are different than the one used by~\cite{prabhakara_high_2023}. Thus, we train two additional models that utilize the previous frame-based approach to compare our model with such a `baseline'. Specifically, we train a model that uses the previous 20 (single-chirp) frames and another model that uses the previous 40 (single-chirp) frames.

\section{Results}
\label{sec:results}
In this section, we present the results from our experimental evaluations, %
comparing our with the baseline models.

\subsection{Performance in Same and Unseen Environments}
\begin{figure}[!t]
\centering
\includegraphics[width=1.00\columnwidth]{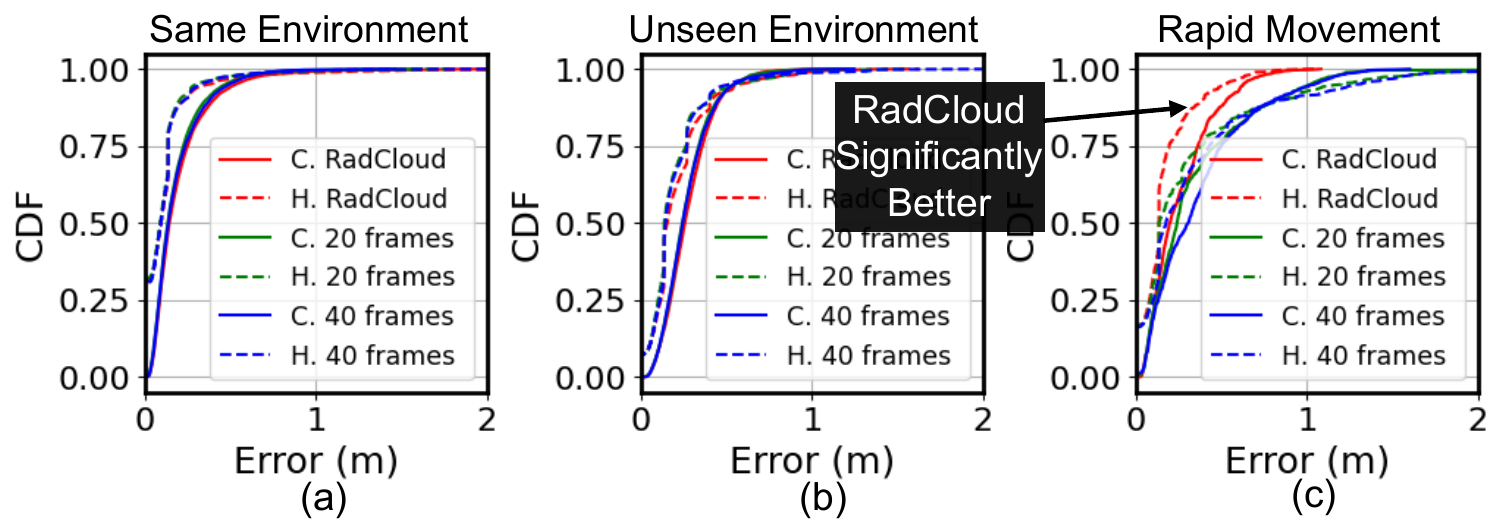} 
\vspace{-16pt}
\caption{Error distributions for (a) same environment, (b) unseen environments, and (c) during rapid movements
}
\vspace{-6pt}
\label{fig:error_comparison}
\end{figure}
\begin{figure}[!t]
\centering
\includegraphics[width=0.998\columnwidth]{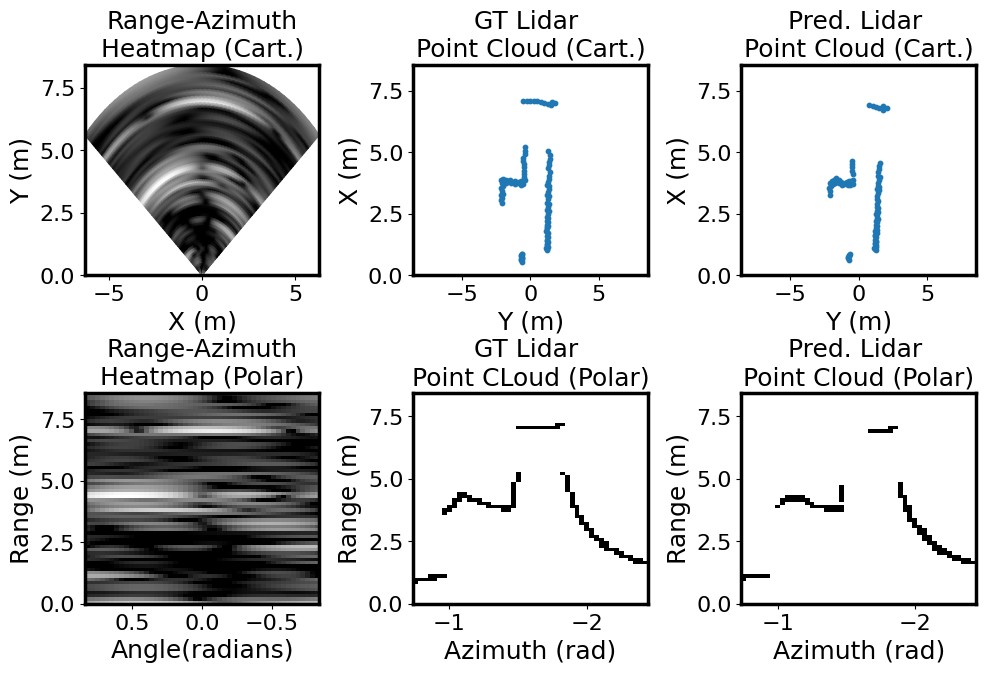} 
\vspace{-16pt}
\caption{Input \radar\ data, ground truth point cloud, and predicted point cloud for nominal operation. The bottom row shows the format at the inputs/outputs of the model. The top row shows the data in 
a Cartesian~format.
}
\vspace{-6pt}
\label{fig:ground_vehicle_result}
\end{figure}

\begin{table}[!t]
\begin{center}
    \caption{Error Comparison - Same Environment}
    \label{table:same_environment_performance_results}
    \vspace{-6pt}
    \begin{tabular}{ 
    m{.24\columnwidth} c| c c c}
    \toprule
    Metric & Units & \ProjectName &  20 frames &  40 frames \\
    \midrule
    Cham. (Mean) & m & 0.20 & \textbf{0.18} & \textbf{0.18}\\
    Cham. (Median) & m & 0.14 & \textbf{0.13} & 0.14\\
    Cham. (90\%) & m & 0.40 & \textbf{0.36} & 0.38\\
    MHaus. (Mean) & m & 0.12 & \textbf{0.11} & \textbf{0.11}\\
    MHaus. (Median) & m & \textbf{0.09} & \textbf{0.09} & \textbf{0.09}\\
    MHaus. (90\%) & m & 0.23 & \textbf{0.20} & 0.23\\
    \bottomrule
    \end{tabular}
\end{center}
\vspace{-4pt}
\end{table}
\begin{table}[!t]
\begin{center}
    \caption{Error Comparison - Unseen Environment}
    \label{table:new_environment_performance_results}
    \vspace{-6pt}
    \begin{tabular}{ 
    m{.24\columnwidth} c| c c c}
    \toprule
    Metric & Units & \ProjectName &  20 frames &  40 frames \\
    \midrule
    Cham. (Mean) & m & 0.27 & 0.27 & \textbf{0.26}\\
    Cham. (Median) & m & 0.26 & \textbf{0.24} & \textbf{0.24}\\
    Cham. (90\%) & m & 0.46 & \textbf{0.45} & \textbf{0.45}\\
    MHaus. (Mean) & m & 0.22 & \textbf{0.19} & 0.20\\
    MHaus. (Median) & m & 0.15 & \textbf{0.09} & 0.14\\
    MHaus. (90\%) & m & 0.41 & \textbf{0.40} & \textbf{0.40}\\
    \bottomrule
    \end{tabular}
\end{center}
\vspace{-4pt}
\end{table}
\fig{\ref{fig:error_comparison}} presents the CDFs of the 
CD and MHD for~the~{\ProjectName} model and the two baseline models, operating in the previously seen (e.g., same as training) environments and unseen environments. \tbl{\ref{table:same_environment_performance_results}} and \tbl{\ref{table:new_environment_performance_results}} summarize the key metrics for each distribution. The results show that our chirp-based approach is nearly as good as the previous frame-based approaches at generating high-resolution point clouds from low-resolution \radar\ data. This is further supported by \fig{\ref{fig:ground_vehicle_result}} showing a predicted point cloud from the {\ProjectName}~model. 

As shown, our model's output is almost identical to the ground truth \lidar\ point cloud, demonstrating that the {\ProjectName} model is well suited for converting low-resolution \radar\ range-azimuth responses to high-resolution \lidar-like 2D point clouds. We also highlight how our model does a good job of capturing complex shapes in the environment like various corner shapes. The results also show that the {\ProjectName} model still generates accurate point clouds even in unseen environments, 
enabling its use on UAVs and UAGs to map or navigate unseen environments. Finally, we highlight the accuracy of our model's predictions with over 90\% of generated point clouds having a CD less than \cm{46} and a MHD less than \cm{41} when compared to the ground truth \lidar\ point cloud, even in unseen environments.
\subsection{Rapid Movement Performance}
\begin{table}[!t]
\begin{center}
    \caption{Error Comparison - Aggressive Maneuvers}
    \label{table:rapid_movement_results}
    \vspace{-6pt}
    \begin{tabular}{ 
    m{.24\columnwidth} c| c c c}
    \toprule
    Metric & Units & \ProjectName &  20 frames &  40 frames \\
    \midrule
    Cham. (Mean) & m & \textbf{0.26} & 0.35 & 0.37\\
    Cham. (Median) & m & \textbf{0.20} & 0.24 & 0.31\\
    Cham. (90\%) & m & \textbf{0.53} & 0.83 & 0.81\\
    MHaus. (Mean) & m & \textbf{0.17} & 0.32 & 0.34\\
    MHaus. (Median) & m & \textbf{0.13} & \textbf{0.13} & 0.17\\
    MHaus. (90\%) & m & \textbf{0.39} & 0.82 & 0.87\\
    \bottomrule
    \end{tabular}
\end{center}
\vspace{-4pt}
\end{table}
\fig{\ref{fig:error_comparison}}(c) presents the CDFs of the CD and MHD for our model and the two baseline models 
during aggressive maneuvers, whereas \tbl{\ref{table:rapid_movement_results}} summarizes the key metrics for each distribution. Compared to 
the other scenarios, 
the performance of the frame-based models noticeably degrades in cases of fast movements. For example, the 20 frames model went from 90\% of predictions having a CD less than \cm{36} and an MHD less than \cm{20}, to 90\% of predictions having a CD less than \cm{83} and an MHD less than \cm{82}. By contrast, \ProjectName\ only experiences slight increases in both CD and MHD while also performing noticeably better than the previous frame-based models. 

\begin{figure}[!t]
\centering
\includegraphics[width=0.98\columnwidth]{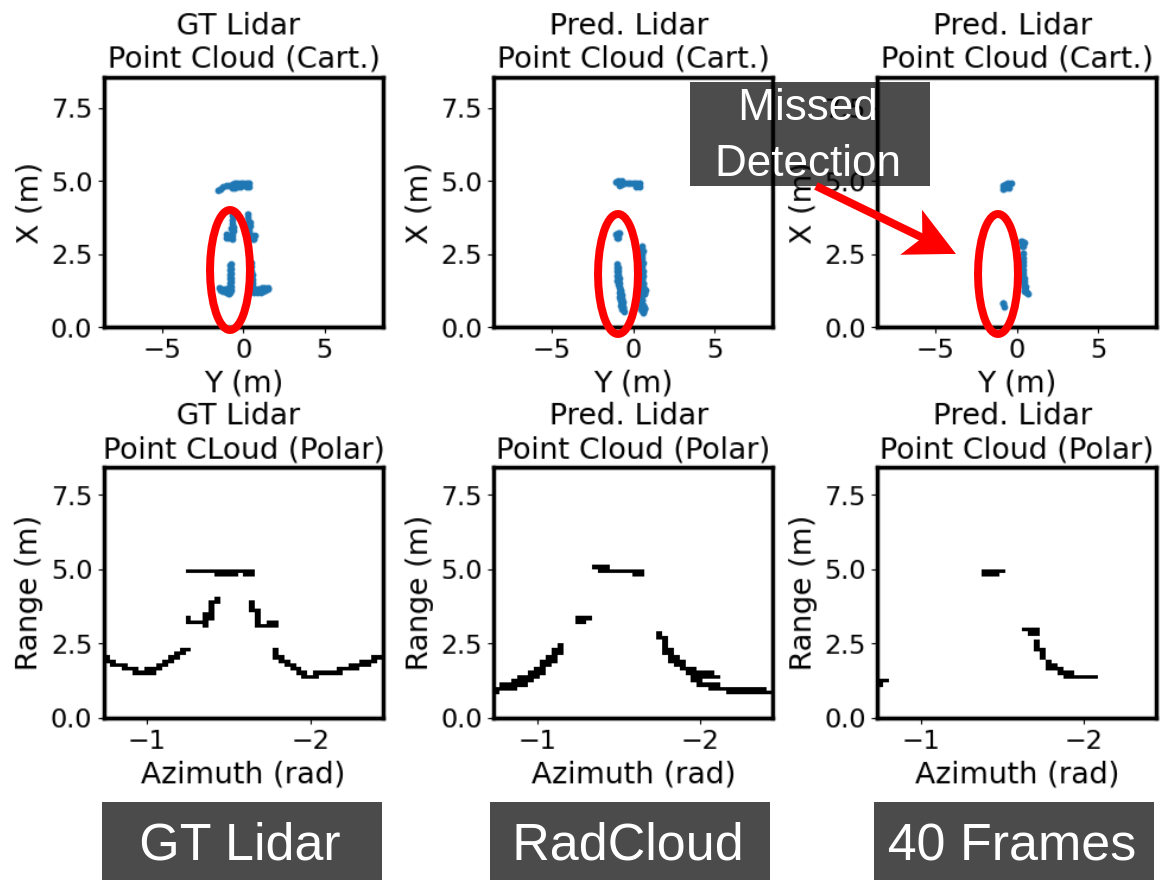} 
\vspace{-2pt}
\caption{Generated point clouds during rapid movements.
}
\vspace{-6pt}
\label{fig:error_comparison_rapid_movement}
\end{figure}

As presented in \fig{\ref{fig:error_comparison_rapid_movement}}, {\ProjectName}'s model still manages to detect the main features of the environment while the 40 frames model fails to detect large features in the environment due to the rapid movements. Overall, these results show that our model is significantly more resilient to aggressive maneuvers compared to the previous frame-based approaches.

\begin{figure}[!t]
\centering
\includegraphics[width=0.998\columnwidth]{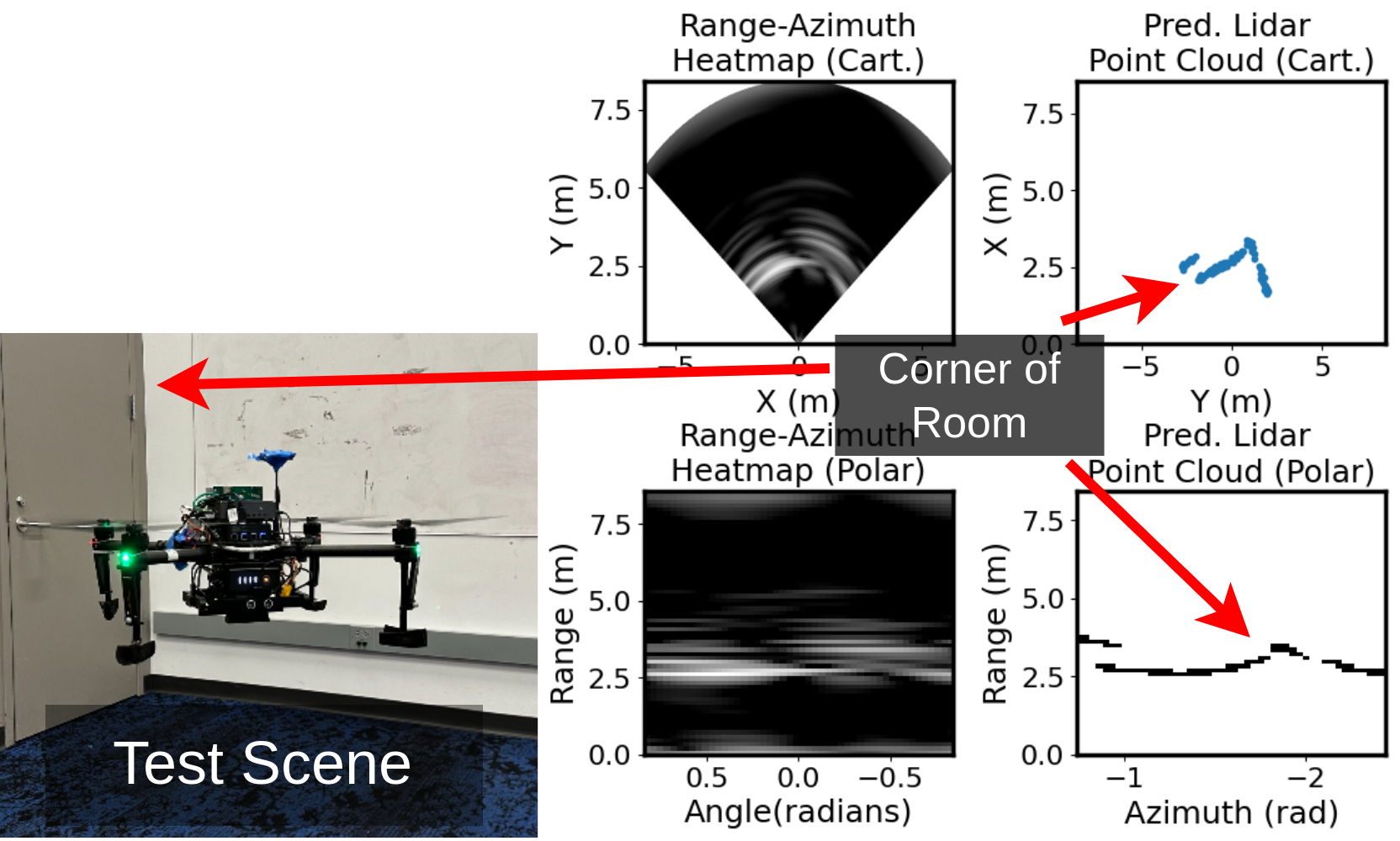} 
\vspace{-14pt}
\caption{Performance on a Drone Based Platform.
}
\vspace{-6pt}
\label{fig:drone_result}
\end{figure}

\vspace{-10pt}
\subsection{UAV Case Study}
\vspace{-2pt}
As discussed in \sect{\ref{sec:experiments}}, we mounted the \radar~and a NUC platform onto a commercially available DJI Matrice 100 drone~\cite{dji_matrice_2023}. While we were not able to obtain ground truth information due to the size and weight of the \lidar~sensor, this case study demonstrates the feasibility and practicality of the {\ProjectName} real-time framework.
We flew the drone inside the environment pictured in \fig{\ref{fig:drone_result}}. Here, we highlight that the model was not trained on this environment nor did we train the model on a drone platform. This is particularly notable as the {\radar} experiences different dynamics (e.g., vibrations) on the drone, and the drone's propellers also add additional noise and interference to the \radar\ data. 

Thus, this case study also demonstrates the {\ProjectName} model's performance in unseen environments and on different platforms. \fig{\ref{fig:drone_result}} presents one of the predicted point clouds generated using our model (a more complete video is available 
at~\cite{RadCloud_Website}).
As shown, 
our system still does a good job of identifying the major features (e.g., walls and corners of the room), even in unseen environments and on different platforms. Combined, these results demonstrate the feasibility and practicality of the {\ProjectName} platform. Thus, we demonstrate \emph{\textbf{a real-time framework for directly converting low resolution {\radar} data to 2D {\lidar}-like point clouds, which can be used for mapping, navigation, and other purposes on UAVs}}.

\section{Conclusion}
\label{sec:conclusion}
In this work, we have presented \ProjectName\, a \emph{real-time} framework for directly deriving high-resolution \lidar-like 2D point clouds from low-resolution \radar\ frames on resource-constrained platforms commonly used in unmanned aerial and ground vehicles; the high-resolution of the point clouds enables their use in accurate environmental mapping, navigation in unknown environments, as well as other robotics tasks. Since existing methods for high-resolution sensing from \radar\ data cannot be used on resource-constrained platforms, \ProjectName~has overcome the challenges presented by these platforms by utilizing a \radar\ configuration with 1/4th the range resolution and deep learning model with $2.25\times$ fewer parameters. Further, we have utilized a novel chirp-based approach making generated point clouds more resilient to aggressive turns, spins, and other rapid movements commonly experienced during UAV and UGV operations. Finally, we have demonstrated the accuracy and applicability of \ProjectName\ on 
commonly used UAVs and UGVs with commercially available \radar\ platforms on board, where we have achieved average frame rates of 15fps even when operating on CPU-only platforms with limited computational~power.

\bibliographystyle{IEEE_styles/IEEEtranMod}
\bibliography{references.bib}

\end{document}